\documentclass{IEEEoj-data}
\usepackage{cite}
\usepackage{amsmath,amssymb,amsfonts}
\usepackage{graphicx,color}
\usepackage{textcomp}
\usepackage{hyperref}
\usepackage{url}
\usepackage[english = american]{csquotes} 
\usepackage{url,multirow,comment,hyperref,amsmath,graphicx,amsmath,subfigure,listings,tabularx,psfrag,wrapfig,cite,verbatim,array, algorithm, algorithmicx, algpseudocode}
\usepackage[
locale = DE 
]{siunitx} 
\usepackage{caption} 
\usepackage{xcolor,mathtools}

\usepackage[colorinlistoftodos]{todonotes}
\presetkeys{todonotes}{fancyline, color=blue!30,size=\tiny}{}

\def\BibTeX{{\rm B\kern-.05em{\sc i\kern-.025em b}\kern-.08em
    T\kern-.1667em\lower.7ex\hbox{E}\kern-.125emX}}
\AtBeginDocument{\definecolor{ojcolor}{cmyk}{0.93,0.59,0.15,0.02}}

\begin{document}
\receiveddate{00 April, 2024}
\reviseddate{11 April, 2024}
\accepteddate{00 April, 2024}
\publisheddate{00 May, 2024}
\currentdate{24 June, 2024}
\doiinfo{DD.2024.0624000}
\title{{\color{black}\textbf{Descriptor:}} 
\textit{Face Detection Dataset for Programmable Threshold-Based Sparse-Vision }}

\author{
Riadul~Islam,~\IEEEmembership{Senior Member,~IEEE},
Sri~Ranga~Sai~Krishna~Tummala,
Joey~Mulé,
Rohith~Kankipati,
Suraj~Jalapally,
Dhandeep~Challagundla,~\IEEEmembership{Student Member,~IEEE},
Chad~Howard,
~and Ryan~Robucci,~\IEEEmembership{Member,~IEEE}

\thanks{R. Islam, S. Tummala, J. Mulé, R. Kankipati, S. Jalapally, D. Challagundla, and R. Robucci are with the Department 
of Computer Science and Electrical Engineering, University of Maryland, Baltimore County, 
MD 21250, USA e-mail: {\{riaduli, stummal1, surajkj1, rohithk1, robucci\}@umbc.edu}.}
\thanks{C. Howard and C. Rizk are with the FRIS Inc. (d/b/a Oculi),
5520 Research Park Drive Catonsville MD, 21228, USA e-mail: {\{chad.howard, charbel.rizk\}@oculi.ai}}
\thanks{This work was supported in part by National Science Foundation (NSF) award number: 2138253, Maryland Industrial Partnerships (MIPS) program award number: MIPS0012, and the UMBC Startup grant.}
\thanks{Copyright (c) 2023 IEEE. Personal use of this material is permitted. 
However, permission to use this material for any other purposes must be 
obtained from the IEEE by sending an email to pubs-permissions@ieee.org.}
}

\markboth{IEEE Transactions on xxxxx}{Shell \MakeLowercase{\textit{et al.}}: ??????}


\begin{abstract}


Smart focal-plane and in-chip image processing has emerged as a crucial technology for vision-enabled embedded systems with 
energy efficiency and privacy.
However, the lack of special datasets providing examples of the data that these neuromorphic sensors compute to convey visual information has hindered the adoption of these promising technologies. Neuromorphic imager variants, including event-based sensors, produce
various representations such as streams of pixel addresses representing time and locations of intensity changes in the focal plane, temporal-difference data, data sifted/thresholded by temporal differences, image data after applying spatial transformations, optical flow data, and/or statistical representations. To address the critical barrier to entry, we provide an annotated, temporal-threshold-based vision dataset specifically designed for face detection tasks derived from the same videos used for Aff-Wild2. By offering multiple threshold levels (e.g., 4, 8, 12, and 16), this dataset allows for comprehensive evaluation and optimization of state-of-the-art neural architectures under varying conditions and settings compared to traditional methods. The accompanying tool flow for generating event data from raw videos further enhances accessibility and usability. We anticipate that this resource will significantly support the development of robust vision systems based on smart sensors that can process based on temporal-difference thresholds, enabling more accurate and efficient object detection and localization and ultimately promoting the broader adoption of low-power, neuromorphic imaging technologies. To support further research, we
publicly released the dataset at \url{https://dx.doi.org/10.21227/bw2e-dj78}.\\

 
{\textcolor{ieeedata}{\abstractheadfont\bfseries{IEEE SOCIETY/COUNCIL}}}     Signal Processing Society (SPS)\\ 
 {\textcolor{ieeedata}{\abstractheadfont\bfseries{DATA DOI/PID}}}     10.21227/bw2e-dj78\\ 
 {\textcolor{ieeedata}{\abstractheadfont\bfseries{DATA TYPE/LOCATION}}}   Images; Maryland, USA

\end{abstract}
\maketitle


\section*{Background}


Algorithm development for smart sensors, those that integrate coincidental sensing and processing, is complicated by a lack of emulation models and challenges in producing synthetic data for testing. The characteristic advantage of focal-plain and on-chip processing for image sensors is that they do not transmit data irrespective of vision tasks.  Instead, such sensors sift and transform signal data into representations of essential information for vision tasks.  Therefore, the average sample-reporting rate is far below the Nyquist Rate, which means a loss of the ability to reconstruct all aspects of the original image by traditional standards.  Furthermore, the parameters of this non-invertible signal pre-processing are often adaptive and controlled algorithmically. This presents the two-fold challenge of developing dependent algorithms for controlling the parameters of sensing and back-end algorithms for computer vision.

State-of-the-art (SOTA) approaches to supervised training for machine learning (ML) algorithms involve a
prolonged iterative approach to the co-determination of salient features from source data along with other processing parameters for
inference. These algorithms rely on unbiased annotation of source data~\cite{Padilla_evaluation:2012, Wenzhen:2016} to guide convergence on parameters of feature extraction and the structures and details of image analysis. However, the key difference is that emulation models for the influence of dynamic parameter selection are lacking which makes exploration challenging.  Even more fundamental is that the underlying characteristic frequency of smart sensors, including those based on ideas from neuromorphic and event-based image sensors~\cite{Amir_dvs:2017, Gallego:2022}, is much higher than conventional sensors. This altogether means that the information presented to back-end algorithms is not readily synthesized from widely available public data.  Approaches to temporal interpolation are sometimes required to synthesize events with a higher temporal resolution than those of the sensors used to produce traditional video datasets. Furthermore, even when existing hardware is accessible for testing, experimental parameter sweeps are non-trivial since the behavior of the subjects of observation, be it a human or cars traveling down the highway, are not typically exactly reproducible nor are the environmental conditions such as illumination. Therefore the goal of the community must be to decouple some of the challenges and provide means to step into algorithm development for these platforms without requiring researchers and developers to first address all of these issues.  In this spirit, we present an annotated dataset representing a practical application of neuromorphic processing to face detection. The application is assumed to involve modest temporal frequencies and would not require temporal interpolation, but a small embedded vision system embodying efficiency and/or privacy would benefit greatly if architected around a smart/event-based camera.    


\subsection*{Motivation}

Event cameras are a promising technology increasingly used for computer vision applications because of their high temporal resolution and low power consumption~\cite{Mueggler:2018}. In contrast to conventional synchronous energy-efficient techniques~\cite{Islam_ncfet:2021, Hyeon:2022, Islam:2018, Park:2020, Islam_sram:2021}, event sensors introduce a paradigm shift by enabling low-power operation through distributed processing. 
However, one of the challenges in developing algorithms for event cameras, such as face detection, is the need for suitable datasets to undertake initial steps in algorithm development. Unlike traditional cameras that capture a sequence of frames that can be annotated later, event cameras capture a series of sparse events~\cite{Guillermo:2019} that could be asynchronously detected or at least processed at a temporal resolution much higher than conventional low-power CMOS camera frame rates. This makes creating datasets for training and testing face detection algorithms difficult. While some datasets exist, like N-Caltech 101, they are typically small and do not represent the faces in real-world scenarios like different poses, illumination, etc.~\cite{Orchard:2015}.
Several other face detection datasets, for example, the WIDER FACE dataset~\cite{Yang_wider:2016}, the CelebA dataset~\cite{Liu_faceattributes:2015}, and the face detection dataset and benchmark (FDDB)~\cite{Jain:2010}. However, none of these datasets provide motion data for faces. 

Existing event-based vision-related benchmark datasets, for example, the N-Caltech101~\cite{Orchard:2015}, N-mnist, Poker-DVS~\cite{Serrano-gotarredona_linares-barranco:2015}, event data for pose, visual odometry, and SLAM~\cite{Elias_dataset:2017} are generated by mounting a dynamic vision sensor (DVS) on a motorized pan-tilt unit and having the sensor move while it views. 
The events are generated by emulated saccades (i.e., capturing events by changing the eye's position) rather than independent object motion in the field of view.
Existing datasets also use a fixed threshold, the difference in pixel intensity between consecutive image frames or the difference between pixel intensity considering a current frame and a reference frame. However, herein, a multi-threshold dataset is presented to facilitate low-bandwidth event-vision. The availability of a threshold ($T_h$) -- based image dataset can enable fine-tuning new neural architectures for performing well on sparse images. Moreover, the $T_h$-based dataset will be fundamentally crucial to designing novel self-adaptive smart vision sensors.

\subsection*{Main Contributions}
The specific contributions of this work are:
\begin{itemize} \renewcommand{\labelitemi}{$\bullet$}
		
	   \item We present a new smart event face dataset (SEFD) of multiple programmable digital thresholds to decouple the challenges of modeling smart sensors and initial algorithm development.
          \item We analyzed pixel activity concerning $T_h$ values to characterize event-based object detection.
	   \item We validate the effectiveness of the proposed dataset through training of industry-standard object detection and localization models.
\end{itemize}

\vspace{-0.250cm}
\label{sec:background}
\subsection*{Existing Event-Based Vision}
Empirically, event-based vision sensors~\cite{Lichtsteiner:2008,Rizk:2012,Joseph_fris:2012,Niwa_isscc:2023,Liu_adaptive:2018,Charbel:2015,Klenk_Enerf:2023,Gallego:2022,Rudnev_cvpr:2023} are promising for high-speed vehicles, robotics, drones, and moving object detection. Optical-flow features are of special interest in bio-inspired approaches, and researchers have developed an algorithm to derive optical flow features from event-based data captured by dynamic vision sensing (DVS) cameras~\cite{Liu_adaptive:2018}. The optical flow-based system employs a block-matching technique~\cite{Liu_iscas:2017} to estimate the flow between sequential DVS event frames. This involves dividing the frames into blocks and finding matching blocks between consecutive frames to determine the motion. 
Another interesting event simulator (ESIM) captures event data using a rendering engine and image sensor trajectory~\cite{Rebecq_esim:2018}. The ESIM uses an image brightness gradient for adaptive sampling.

Other researchers have facilitated DVS using the v2e toolbox~\cite{Hu_v2e:2021}. This toolbox employs various parameters, including temporal noise, leak events, pixel intensity, and the Gaussian threshold method. While proficient in generating event data, it is noteworthy that the v2e toolbox lacks explicit guidelines for creating event-based facial data, a distinctive feature of our proposed approach. 
Another biological retina-based tool, namely RetinoSim, was designed to synthesize event-based data for exploring neuromorphic vision architectures~\cite{Sengupta:2022}. 
It addresses the need for realistic and diverse event-based data to develop and evaluate such architectures effectively. In a hybrid approach, researchers used ``frame of events" to combine conventional frame scanning and DVS approaches to improve computational latency and memory consumptions~\cite{Venkatachalam:2023}.

\vspace{-0.250cm}
\subsection*{Existing Neural Network Architectures for Image Classification and Localization}
The task of object detection and localization within the domain of event-vision presents a non-trivial challenge, primarily stemming from the absence of salient feature representations within the input data. In recent years, you only
look once (YOLO) creates great attention from researchers for object detection and localization~\cite{Redmon_yolo:2016}.
The initial version of the YOLO model uses 24 convolutional layers and two fully-connected layers. 
This approach divides an image into $S \times S$ grids and detects an object based 
on the location of the center of an object that falls into the grid. To improve the performance, YOLO-v4~\cite{Bochkovskiy_yolov4:2020} uses weighted residual connections (WRC) and
cross-stage partial connections (CSPs).

In this work, we benchmarked our dataset using YOLO-v4~\cite{Bochkovskiy_yolov4:2020} and YOLO-v7~\cite{Wang_yolov7:2022}. YOLO-v4 consists of backbone, neck, and head networks. 
When YOLO-v4 targets GPU, the backbone network uses CSP Network (i.e., CSPDarknet53)~\cite{Wang_cspnet:2020}. 
YOLO-v7 introduces an updated and optimized architecture compared to previous YOLO versions, with the use of Extended Efficient Layer Aggregation Network (E-ELAN) for better feature map integration compared to YOLO-v4's PANet and SPP.

Google Brain research team proposed an interesting architecture for object detection and localization called EfficientDet-b0~\cite{Tan_efficientdet:2020}. 
Another set of interesting convolutional neural network (CNN)-based architectures proposed by the Google research team, especially suitable for embedded and mobile vision applications, called MobileNets-v1~\cite{Howard_mobilenets:2017}. 
Both EfficientDet-b0 and MobileNets-v1 will be used in this research for benchmarking.
\vspace{-0.250cm}

\section*{Collection Methods and Design}

\label{sec:proposed_method}

\subsection*{Description of Source Dataset}
\label{sec:aff_wild}
In this work, we used the Aff-Wild2 dataset as input source videos~\cite{Kollias_affw2:2018}.
The Aff-Wild2 dataset features emotional descriptors described in terms of valence and arousal. Valence indicates the intensity of positive and negative emotions, while the latter suggests the power of triggering an emotion. The dataset contains 298 videos of 200 subjects (about 15 hours of data) annotated by seven lay experts considering valence and arousal, all captured in a natural state without any external stimulation. 

\begin{figure}[h!]
\begin{center}
\includegraphics[width = 0.4\textwidth]{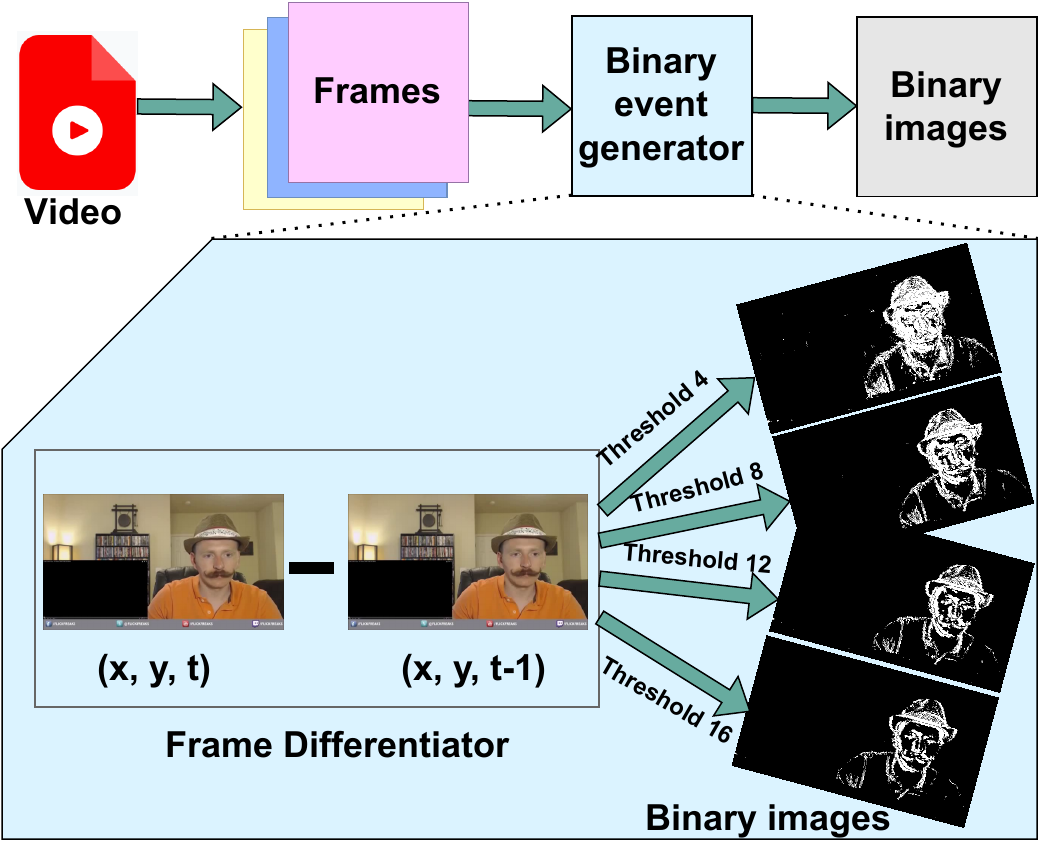}
\vspace{-0.250cm}
\end{center}
\caption[ ]
{The proposed tool flow converts raw video data into image frames, and then the binary event generator differentiates temporally different image frames and a set of reference threshold values to generate thresholded images.}
\label{fig:bin_img_gen_flow}
\vspace{-0.50cm}
\end{figure}
\vspace{-0.250cm}
\subsection*{Proposed Multi-Threshold Image Generation Tool Flow}
\label{sec:multi_thres_gen}
The existing face detection benchmark datasets do not provide motion data to capture pixel activity~\cite{Liu_faceattributes:2015, Yang_wider:2016, Jain:2010}. Our research resolves this issue with our proposed smart event generation tool flow and datasets. The proposed binary image generation framework is shown in Figure~\ref{fig:bin_img_gen_flow}.
In this tool flow, we used an input video file to generate image frames considering the frame rate or temporal difference of frames. Then, we used our proposed binary event generator to create binary image frames considering a set of threshold values, as shown in Figure~\ref{fig:bin_img_gen_flow}.

The proposed threshold-based smart event generation methodology is shown in Algorithm~\ref{alg:th_event_frame_gen}. The algorithm uses a video file ($Vi$), the threshold value ($T_h$), and a user-defined frame temporal difference ($T_p$) as inputs. First, the algorithm generates image frames using the $GenFrame()$ function considering frame temporal differences in Line~\ref{alg1:line4}. It recursively derives the previous frame from the current frame to compute the differential frame from Line~\ref{alg1:line5} to Line~\ref{alg1:line8}. Depending on the threshold value, we update the differential frame pixel intensity from Line~\ref{alg1:line9} to Line~\ref{alg1:line16}. Finally, the algorithm returns the binary frames in Line~\ref{alg1:return}.

\vspace{-0.40cm}
\begin{algorithm}

        \caption{$T_h$-based smart event generation algorithm}
        \label{alg:th_event_frame_gen}
        \begin{algorithmic}[1]
            \State {\bf Input:} $Vi$, input video; $T_h$, threshold; $T_p$, temporal difference
            \State {\bf Output:} Sparse video, videos with frames featuring motion smart events
            \State $Frames = \{frame_0, frame_1,\dots,frame_n\} = GenFrame(Vi, T_p)$ \Comment{Generate image frames from the input video file.}  \label{alg1:line4}
            
            \ForAll {$frame_{i+1} \in Frames$} \label{alg1:line5}
                \State $CurrFrame = frame_{i+1}$ \Comment{Initialize current frame.} \label{alg1:line6}
                \State $PrevFrame = frame_i$ \Comment{Initialize previous frame.} \label{alg1:line7}
                \State $DiffFrame_i = \{px_{(0,1)},\dots,px_{(M,N)}\} = CurrFrame - PrevFrame$ \Comment{Compute a differential frame considering an image size of $M \times N$.} \label{alg1:line8}
                \ForAll {$px_{(x,y)} \in DiffFrame_i$} \label{alg1:line9}
                    \If {$px_{(x,y)} > T_h$}  \label{alg2:line10}
                        \State $px_{(x,y)} = 1$ \Comment{Detecting pixel activity} \label{alg2:line11}
                    \Else \label{alg2:line12}
                        \State $px_{(x,y)} = 0$ \label{alg2:line13}
                    \EndIf \label{alg2:line14}
                \EndFor \label{alg1:line15}
            \EndFor \label{alg1:line16}
            \State \Return $DiffFrames = \{DiffFrame_1, DiffFrame_2,\dots,$ $DiffFrame_{n-1}\}$ \Comment{Return differential frames.} \label{alg1:return}
        \end{algorithmic}
        
\end{algorithm}

\subsection*{Image Annotation Guideline}
\label{sec:img_anno}
For object detection and localization, building a bounding box (BBox) of a rectangular portion of an object is standard practice. We used four lay expert human annotators to draw rectangular boxes around objects of interest in an image (i.e., face) to provide training data for object detection and recognition algorithms.

\begin{figure}[t]
\begin{center}
\includegraphics[width = 0.5\textwidth]{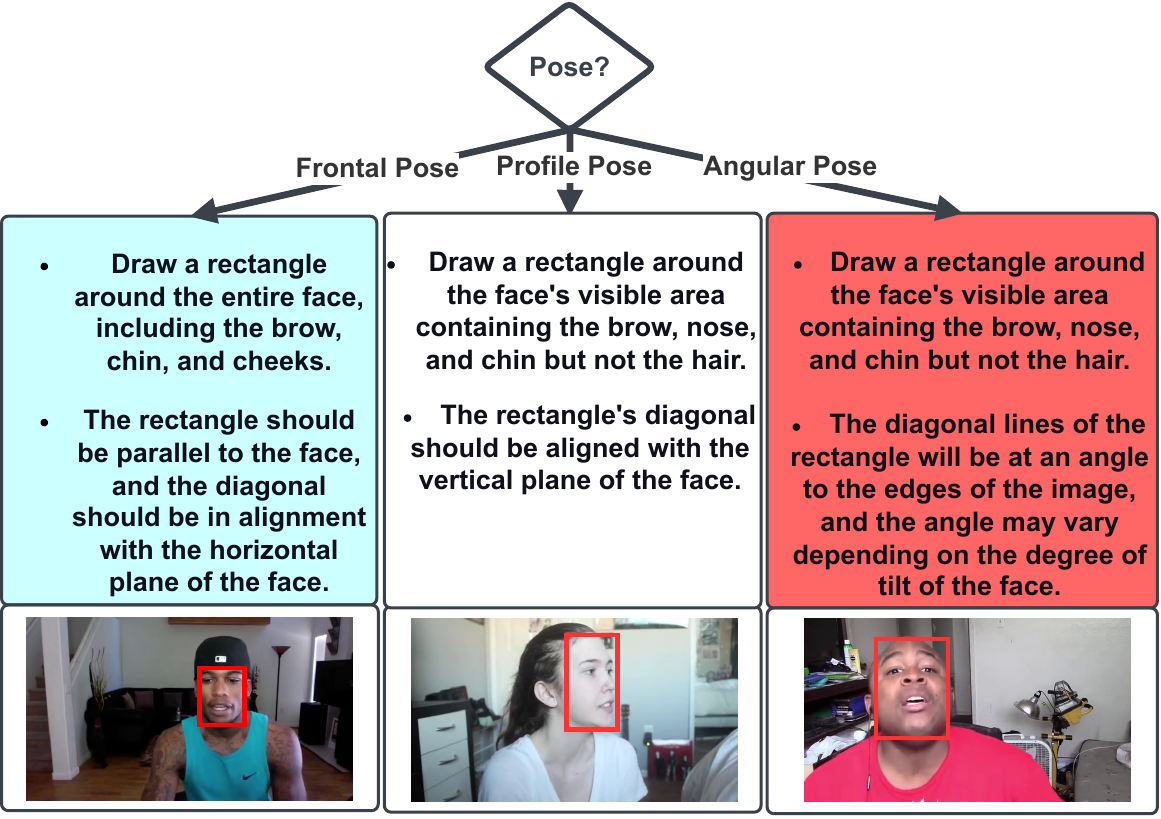}
\end{center}
\caption[ ]
{Face annotation involved creating a rectangle over the face, with the rectangle's orientation varying by pose: parallel and horizontally aligned for frontal, vertically aligned for profile, and angled for angular, depending on the face's tilt.
}
\label{fig:face_anno}
\vspace{-0.450cm}
\end{figure}

\begin{table}[t!] \scriptsize
    \vspace{-0.25cm}
    \renewcommand{\arraystretch}{1.5}
    \caption{The proposed tool flow used audio video interleave (AVI) format video length ranging from 0.1 min to 14.47 min and converted to (threshold = 4, 8, 12, and 16) image frames of portable network graphic (PNG) format.
    \label{tab:data_attr}}
    \centering
    \scalebox{0.7}{ 
        \begin{tabular}{|c|c|c|c|c|c|}
            \hline
            \multirow{2}{*}{Attributes} & \multirow{2}{*}{Video Descriptions} & \multicolumn{4}{c|}{Image Descriptions} \\
            \cline{3-6}
              &   & $T_h = 4$ & $T_h = 8$ & $T_h = 12$ & $T_h = 16$ \\
            \hline
            Length of videos & 0.10--14.47 min & -- & -- & -- & -- \\
            \hline
            Video format & AVI & PNG & PNG & PNG & PNG \\
            \hline
            Avg. image resolution (AIR) & $416 \times 416$ & $416 \times 416$ & $416 \times 416$ & $416 \times 416$ & $416 \times 416$ \\
            \hline
            Standard deviation of AIR & 0 & 0 & 0 & 0 & 0 \\
            \hline
            Median image resolution & $416 \times 416$ & $416 \times 416$ & $416 \times 416$ & $416 \times 416$ & $416 \times 416$ \\
            \hline
            \# of Videos/ Images & 100 & 9130 & 9130 & 9130 & 9130 \\
            \hline
        \end{tabular}
    }
    \vspace{-0.25cm}
\end{table}

For consistency in the process of annotating facial regions with bounding
rectangles, a set of explicit guidelines was used. These guidelines detail the procedure for incorporating facial landmarks to encapsulate the face within the rectangle effectively.
It is of paramount importance, under all circumstances, to ensure that the bounding rectangle maintains a high degree of precision and does not deviate from an appropriate size range, avoiding both excessive enlargement and undue reduction. The size and placement of the rectangle, throughout the duration of the video or image sequence, should remain consistent with the facial features, thereby upholding a standardized and coherent representation of the face. 

The dataset consists of three basic poses (i.e., frontal, profile, and angular), as shown in Figure~\ref{fig:face_anno}. For the frontal pose, the annotators draw a rectangle around the entire face, including the brow, chin, and cheeks. The rectangle should
be parallel to the face, and the diagonal should be in alignment with the horizontal plane of the face. For the profile pose, a rectangle was drawn around the face's visible area containing the brow, nose, and chin but not the hair. The rectangle's diagonal should be aligned with the vertical plane of the face. Finally, for the angular pose, the annotator drew a rectangle around
the face's visible area containing the brow, nose, and chin but not the hair. The diagonal lines of the
rectangle will be at an angle to the edges of the image, and the angle may vary depending on the degree of tilt of the face, as shown in Figure~\ref{fig:face_anno}.

The data source is the original data from the Aff-Wild2 database. The video lengths ranged from 0.1 min to 14.47 min~\cite{Kollias_affw2:2018}, as shown in Table~\ref{tab:data_attr}. The image data was resized so that the average image resolution (AIR) was converted from the original $607\times 359$ to $416\times 416$. On average 100 images
were extracted from each video resulting in 10k full-frame images.  From each of these these $\sim 9.1k$ images were produced considering each threshold (i.e., 4, 8, 12, and 16) using Algorithm~\ref{alg:th_event_frame_gen}, resulting in additional $\sim 36.5k$ images.

\section*{Validation and Quality}
We implement the proposed threshold-based smart event generation Algorithm~\ref{alg:th_event_frame_gen} using Python programming 
language. All the computation was performed on an Intel Xeon Gold processor with 64 GB RAM, equipped with an NVIDIA
Quadro P4000 GPU using Ubuntu 22.04. We used GPU to train, validate, and test SOTA neural network architectures using the proposed dataset. The Algorithm~\ref{alg:th_event_frame_gen} used Aff-Wild2 video dataset~\cite{Kollias_affw2:2018} for threshold-based event-frame generation.
For analysis, we used widely used anchor boxes-based highly accurate YOLO-v4 and YOLO-v7 models. We also used EfficientDet-b0~\cite{Tan_efficientdet:2020}, which is designed to be a lightweight yet effective model for object detection tasks.
In addition, we used MobileNets-v1~\cite{Howard_mobilenets:2017}, emphasizing computational efficiency and adaptability, which is generally used in scenarios where real-time, on-device object detection and localization are required. 

\begin{figure}[t!]
\begin{center}
\includegraphics[width = 0.4\textwidth]{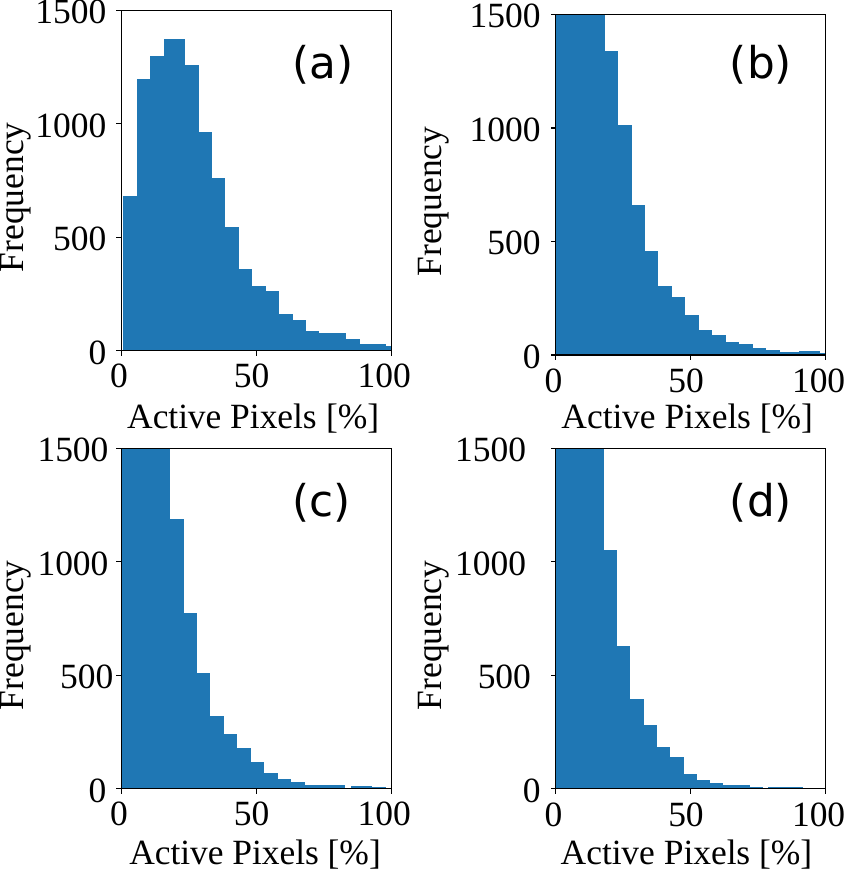}
\vspace{-0.250cm}
\end{center}
\caption[ ]
{For a 333 ms frame time difference, lower thresholds yield higher average active pixel rates, with (a) $T_h = 4$ yielding 25.39\% activity and (b) $T_h = 8$ yielding 17.86\%, (c) $T_h = 12$ yielding 14.27\% and (d) $T_h = 16$ yielding 11.94\%.
    }
\label{fig:pixel_act_tem10}
\vspace{-0.50cm}
\end{figure}

\vspace{-0.10cm}
\subsection*{Pixel Activity Computation}

In general, event-based vision sensing is based on {\emph{pixel activity}}--- a term used herein to refer to pixels associated with a change in sensed light intensity that surpasses a given $T_h$ and triggers communication. The sensors detect spatial motion of objects through the change of pixel intensities. \emph{Pixel activity} shows which pixels detect motion at various thresholds. Analyzing pixel activity allows for the optimization of threshold parameters, enhancing accuracy and efficiency in event detection. Hence, we performed analysis on the \emph{pixel activity} of the proposed dataset considering both the temporal and the intensity difference as parameters.

In order to compute the {\emph{pixel activity}}, we considered intensity threshold values of 4, 8, 12, and 16. Figure~\ref{fig:pixel_act_tem10} shows the analysis regarding the temporal difference of frames of 333 ms. According to our analysis on $\sim 10k$ images, each image contains 25.39\% active pixels on average, considering a $T_h$ of 4, as shown in Figure~\ref{fig:pixel_act_tem10}(a). For $T_h$ of 8, each image contains 17.86\% active pixels on average, as shown in Figure~\ref{fig:pixel_act_tem10}(b). Using $\sim 10k$ images and considering a $T_h$ of 12, we observed 14.27\% of average {\emph{pixel activity}}, as shown in Figure~\ref{fig:pixel_act_tem10}(c). Using the same amount of images, when we considered a $T_h$ value of 16, we observed 11.94\% average {\emph{pixel activity}}, as shown in Figure~\ref{fig:pixel_act_tem10}(d).

Besides, we computed the {\emph{pixel activity}} inside the BBox, which is critically important for object detection and localization. For a $T_h$ value of 16, the average number of active pixels in the BBox is 1.13\%. As expected, by decreasing the intensity $T_h$ value, the average {\emph{pixel activity}} in the bounding boxes increased. When considering $T_h$ values of 8 and 4, the average active pixels in bounding boxes are 1.68\% and 2.34\%, respectively.

\begin{figure}[t!]
\begin{center}
\includegraphics[width = 0.45\textwidth]{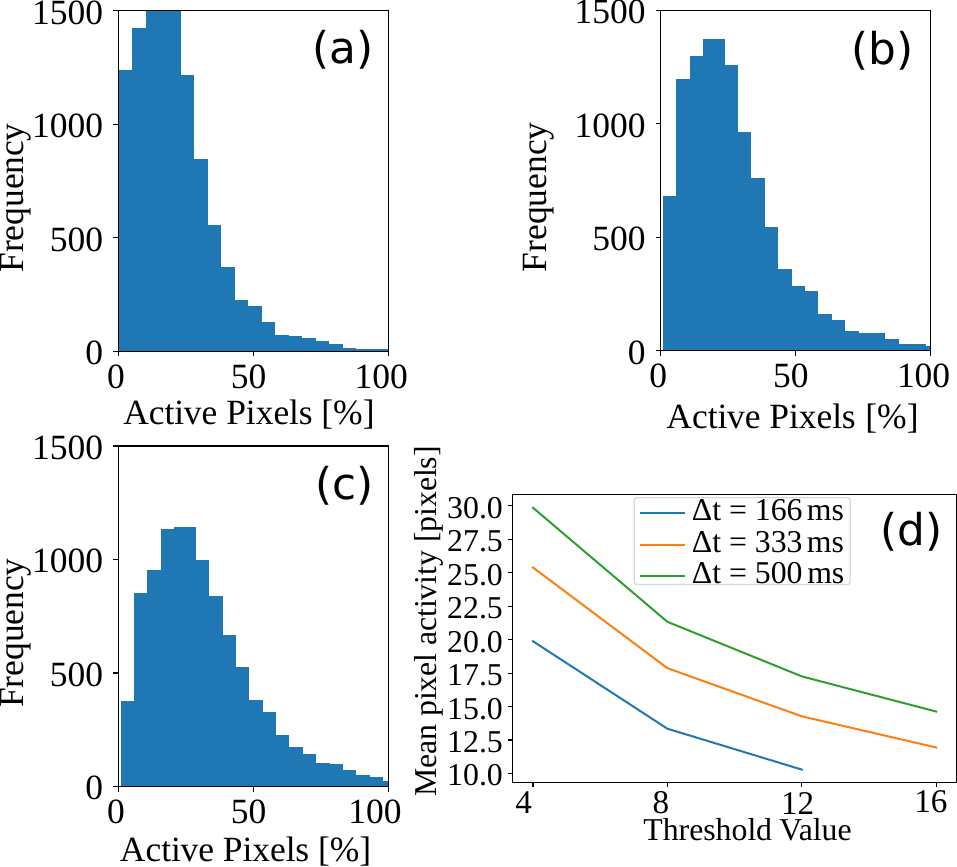}
\vspace{-0.250cm}
\end{center}
\caption[ ]
{For a constant $T_h$ of 4, the average active pixel count increases with the larger interframe time difference, (a) 19.88\% average pixel activity at 166 ms, (b) 25.40\% at 333 ms, and (c) 29.86\% at 500 ms. Conversely, (d) the average active pixel count decreases as $T_h$ increases.
}
\label{fig:pixel_act_fr_var}
\vspace{-0.80cm}
\end{figure}
In addition, we traced the pixel activities due to variations in the frame's temporal differences. The average {\emph{pixel activity}} is 19.88\%, 25.40\%, and 29.86\% considering temporal frame difference of 5 (i.e., 166 ms), 10 (i.e., 333 ms), and 15 (i.e., 500 ms), as shown in Figure~\ref{fig:pixel_act_fr_var}(a),  Figure~\ref{fig:pixel_act_fr_var}(b), and Figure~\ref{fig:pixel_act_fr_var}(c), respectively. Figure~\ref{fig:pixel_act_fr_var}(d) exhibits the overall {\emph{pixel activity}} considering the variation of $T_h$ values and image frames temporal differences. Clearly, there is a positive correlation between temporal difference and {\emph{pixel activity}}, which implies that when the time delay between two frames rises, so does pixel activation. 
The $T_h$, on the other hand, has an anticorrelation with {\emph{pixel activity}}. {\emph{Pixel activity}} reduces as the $T_h$ value increases, which indicates that an imager with a higher $T_h$ would produce fewer pixel activations per frame on average. In contrast, a lower $T_h$ would create more pixel activations per frame. 

\vspace{-0.50cm}

\section{Records and Storage}

\begin{figure}[ht!]
\begin{center}
\includegraphics[width = 0.45\textwidth]{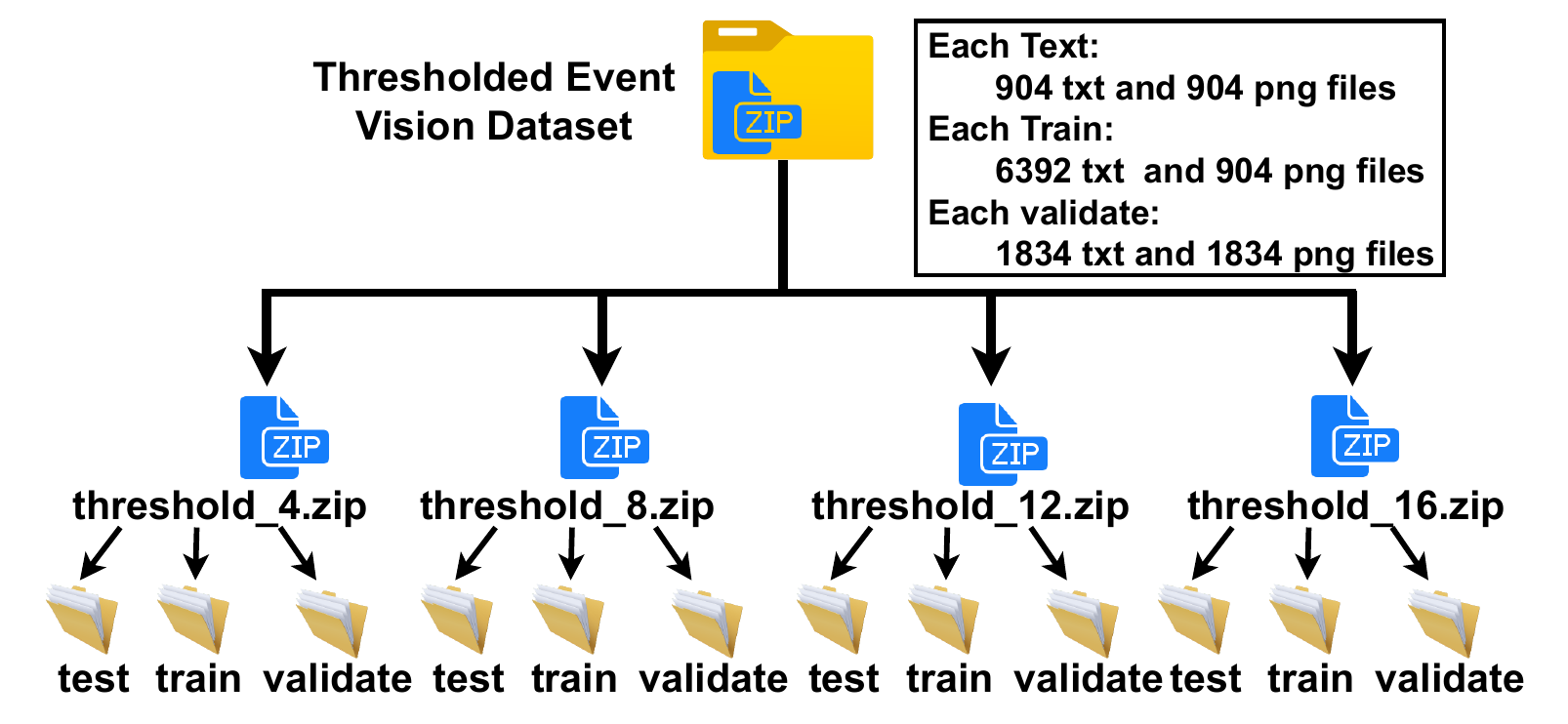}
\vspace{-0.250cm}
\end{center}
\caption[ ]
{Outline of dataset directory structure
}
\label{fig:directories}
\vspace{-0.80cm}
\end{figure}

The dataset is organized into multiple ZIP files (Figure~\ref{fig:directories}), each representing a different threshold level, specifically named threshold\_4.zip, threshold\_8.zip, threshold\_12.zip, and threshold\_16.zip. Within each ZIP file, the data is structured into three sub-directories: test, train, and validate, following common conventions for NN implementations. Each sub-directory contains a set of text (annotation) and image files. The test, train, and validate sub-directories hold 904 text and 904 image files, 6392 text and 6392 image files, and 1834 text and 1834 image files, respectively. 

\section*{Insights and Notes}
\label{sec:analysis}

\subsection*{Benchmarking with SOTA NN Architectures}

\begin{figure*}[h!]
\begin{center}
\includegraphics[width = 0.78\textwidth]{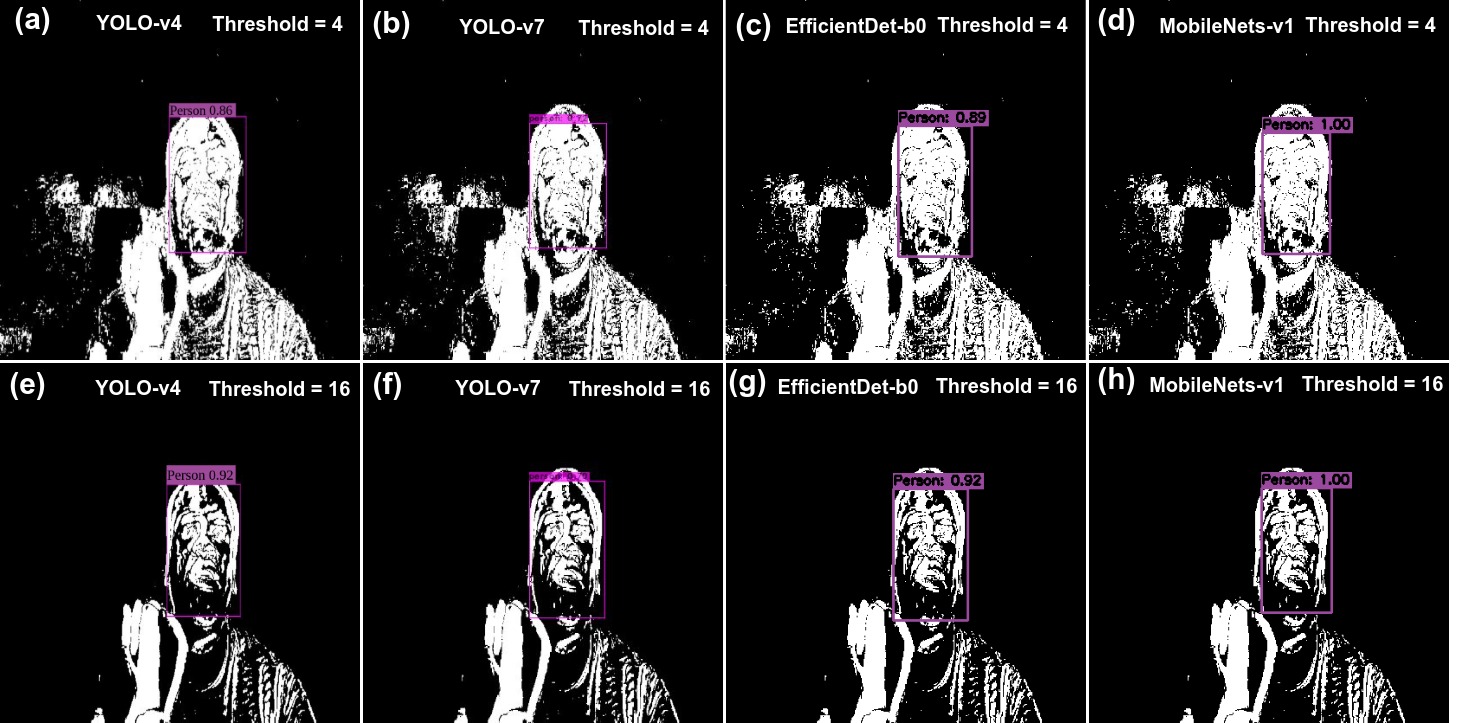}
\vspace{-0.250cm}
\end{center}
\caption[ ]
{A comprehensive assessment of the $T_h$-based dataset using SOTA neural networks demonstrated efficient object detection and localization. (a-d) Models YOLO-v4, YOLO-v7, EfficientDet-b0, and MobileNets-v1 applied to $T_h = 4$ images showed high detection performance, with YOLO-v7 achieving a confidence level of 94.1\%. (e-h) For $T_h = 16$ images, all models maintained strong performance, although YOLO-v7 exhibited a slight drop in confidence level to 91.7\%.

}
\label{fig:class_loca_benchmarks}
\vspace{-0.70cm}
\end{figure*}
In order to enable event-vision using the proposed event-based face dataset, we used SOTA neural network architectures.
Among different existing architectures, we used large YOLO-v4 model~\cite{Wang_cspnet:2020}, YOLO-v7~\cite{Wang_yolov7:2022}, and two smaller sized high-performance models, EfficientDet-b0~\cite{Tan_efficientdet:2020} and MobileNets-v1~\cite{Howard_mobilenets:2017}. In this analysis, we used all four proposed thresholds (i.e., 4, 8, 12, and 16) datasets, as shown in Table~\ref{tab:benchmarks}. We measured the performance of each neural network architecture considering Precision, Recall, F1-score, intersection over union (IOU), average Precision at 50\% IoU (AP50), and average Precision at 70\% IoU (AP75). The results demonstrate evidence of object detection with a confidence level of 86\% for $T_h = 4$ and 92\% for $T_h = 16$ when utilizing the YOLO-v4 model, as presented in Figure~\ref{fig:class_loca_benchmarks}(a) and Figure~\ref{fig:class_loca_benchmarks}(e), respectively. Predictively, using $T_h = 4$ dataset, the YOLO-v7 model has a better average confidence level of 94.1\% compared to $T_h = 16$ dataset, which has a confidence level of 91.7\%, as shown in Figure~\ref{fig:class_loca_benchmarks}(b) and Figure~\ref{fig:class_loca_benchmarks}(f), respectively. Meanwhile, Figure~\ref{fig:class_loca_benchmarks}(c) and Figure~\ref{fig:class_loca_benchmarks}(g) showcase the outcomes of object detection with confidence levels of 89\% for $T_h = 4$ and 92\% for $T_h = 16$, and localization using the EfficientDet-b0. 

Remarkably, for the same sample, the MobileNets-v1 model exhibited superior performance, achieving object detection with a confidence level of 100\% for $T_h$ = 4 and 100\% for $T_h$ = 16, and successfully localizing the object, as illustrated in Figure~\ref{fig:class_loca_benchmarks}(d) and Figure~\ref{fig:class_loca_benchmarks}(h), respectively.

\begin{figure*}[h!]
\begin{center}
\includegraphics[width = 0.8\textwidth]{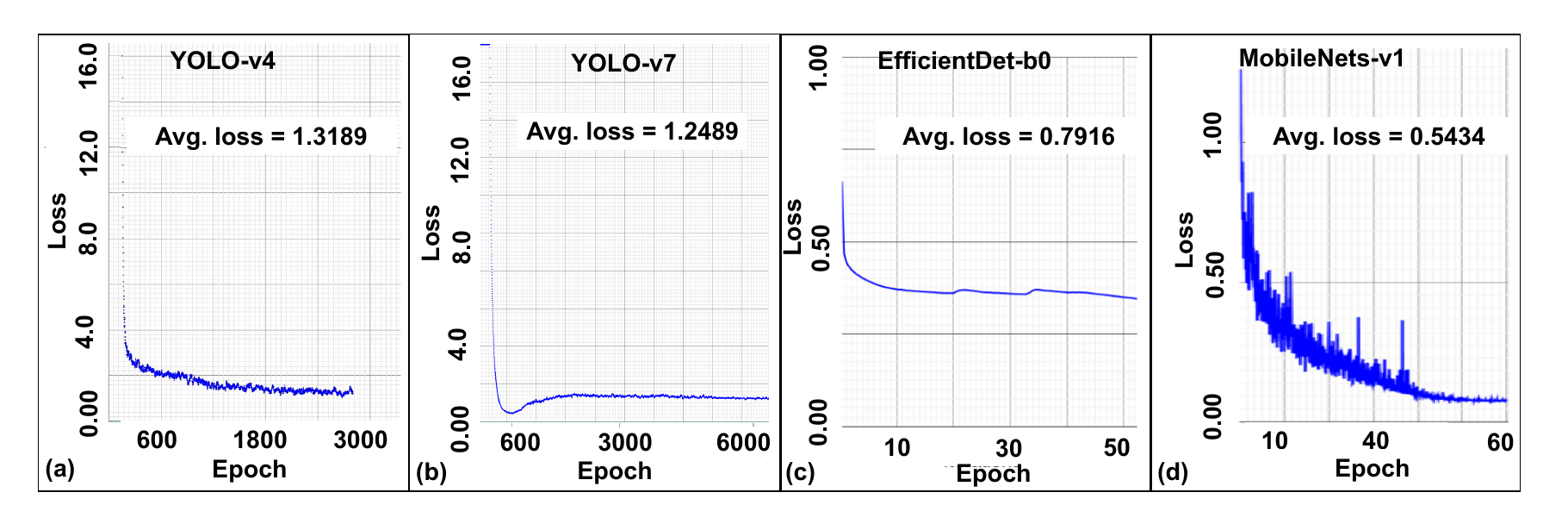}
\end{center}
\vspace{-0.250cm}
\caption[ ]
{All benchmark models achieved low training losses: (a) YOLO-v4 reached an average loss of 1.3189 after 3,000 epochs, (b) YOLO-v7 recorded an average loss of 1.2489 after 6,000 epochs, (c) EfficientDet-b0 converged quickly, achieving an average loss of 0.7916 within 50 epochs, and (d) MobileNets-v1 achieved the lowest average loss of 0.5434 in just 60 epochs.
}
\label{fig:loss_benchmarks}
\vspace{-0.50cm}
\end{figure*}

For the YOLO-v4 and YOLO-v7 models, we used 6359 images of each threshold dataset.
Once trained, we used 915 images from each of the different $T_h$ value datasets for testing. The  YOLO-v4 and YOLO-v7 models have an average training loss of 1.3189 and 1.2489, as shown in Figure~\ref{fig:loss_benchmarks}(a) and Figure~\ref{fig:loss_benchmarks}(b), respectively. 

Considering all four proposed $T_h$ values, the average test Precision, Recall, and F1-score are 94.30\%, 90.50\%, and 92.50\%, respectively. Besides, the YOLO-v4 model achieves, on average, 72.54\%, 92.96\%, and 40.17\% IoU, AP50, and AP75, respectively, considering equal amounts of test data for each kind of $T_h$, as shown in Table~\ref{tab:benchmarks}. The YOLO-v7 has similar test Precision, Recall, F1-score, IoU, and AP50; however, 32.96\% better AP75 compared to the YOLO-v4. 

Like YOLO models, the EfficientDet-b0 and the MobileNets-v1 models used all four threshold-valued datasets for training and testing. The EfficientDet-b0 and the MobileNets-v1 models have an average training loss of 0.7916 (Figure~\ref{fig:loss_benchmarks}(c)) and 0.5434 (Figure~\ref{fig:loss_benchmarks}(d).

Considering all four threshold values, the EfficientDet-b0 architecture achieved an average test Precision, Recall, and F1-score are 97.70\%, 99.37\%, and 98.53\%, respectively. In addition, the EfficientDet-b0 model achieves, on average, 79.93\%, 97.70\%, and 75.45\% IoU, AP50, and AP75, respectively, considering equal amounts of test data for each kind of $T_h$, as shown in Table~\ref{tab:benchmarks}. Among all the models, MobileNets-v1 has highest average Precision, Recall, F1-score, and IoU of 96.30\%, 98.19\%, 97.24\%, and 81.24\%, respectively.

\begin{table*}[h]
\vspace{-0.25cm}
	\renewcommand{\arraystretch}{1.0}
	\caption{
		The YOLO-v4 model exhibits a parameter count that is notably greater, approximately $15.46 \times$ more than that of the EfficientDet-b0 model and about $9.97 \times$ more than that of the MobileNets-v1 model; furthermore, in the context of the proposed $T_h$ = 8 datasets, the MobileNets-v1 model attains the highest IoU of 82.53\% among the models considered, and the EfficientDet-b0 model demonstrates the highest average Precision at 50\% (AP50) of 98.08\% within the same $T_h$ = 8 valued dataset, surpassing the performance of other architectures and datasets with different thresholds.
		\label{tab:benchmarks}}
    \centering
    \vspace{-0.150cm}
    \begin{tabular}{|c|c|c|c|c|c|c|c|c|c|c|c|c|}
    \hline
        \multirow{2}{*}{ Models} &\multirow{2}{*}{\# of Parameters} & \multirow{2}{*}{$T_h$} & \multicolumn{2}{c|}{Dataset split}  &  \multicolumn{3}{c|}{Metrics} &  \multirow{2}{*}{IoU (\%)} & \multirow{2}{*}{AP50} & \multirow{2}{*}{AP75} \tabularnewline
			\cline{4-8}
        ~ & ~ & ~ & Training & Testing  & Precision & Recall & F1-score & ~ & ~ & ~  \\ \hline
        \multirow{4}{*}{YOLO-v4~\cite{Bochkovskiy_yolov4:2020}} &  \multirow{4}{*}{60.3M} & 4 & 6392 & 904   & 94.00 & 89.00 & 91.00 & 72.77 & 90.93 & 45.85 \tabularnewline 
			\cline{3-11}
        ~ & ~ & 8 & 6392 & 904   & 89.00 & 89.00 & 89.00 & 69.04 & 90.37 & 44.57  \tabularnewline 
			\cline{3-11}
        ~ & ~ & 12 & 6392 & 904  & 91.00 & 86.00 & 89.00 & 70.57 & 90.32 & 43.89 \tabularnewline 
			\cline{3-11}
        ~ & ~ & 16 & 6392 & 904  & 93.00 & 79.00 & 85.00 & 71.36 & 86.54 & 38.74 \\ \hline
         \multirow{4}{*}{YOLO-v7~\cite{Wang_yolov7:2022}} &  \multirow{4}{*}{25.2M} & 4 & 6392 & 904   & 94.00 & 92.00 & 93.00 & 75.25 & 94.55 & 60.80 \tabularnewline 
			\cline{3-11}
        ~ & ~ & 8 & 6392 & 904   & 92.00 & 90.00 & 91.00 & 73.62 & 91.82 & 64.31  \tabularnewline 
			\cline{3-11}
        ~ & ~ & 12 & 6392 & 904  & 95.00 & 87.00 & 91.00 & 76.40 & 92.45 & 58.12  \tabularnewline 
			\cline{3-11}
        ~ & ~ & 16 & 6392 & 904  & 95.00 & 95.00 & 90.00 & 76.11 & 90.73 & 56.45 \\ \hline
         \multirow{4}{*}{EfficientDet-b0~\cite{Tan_efficientdet:2020}} &  \multirow{4}{*}{3.9M} & 4 & 6392 & 904  & 97.78 & 99.34 & 98.56 & 80.45 & 97.78 & 76.38  \tabularnewline 
			\cline{3-11}
        ~ & ~ & 8 & 6392 & 904  & 98.08 & 99.45 & 98.76 & 80.53 & 98.08 & 78.77 \tabularnewline   
			\cline{3-11}
        ~ & ~ & 12 & 6392 & 904 & 97.63 & 99.34 & 98.48 & 80.06 & 97.63 & 75.72  \tabularnewline 
			\cline{3-11}
        ~ & ~ & 16 & 6392 & 904 & 97.29 & 99.34 & 98.30 & 78.69 & 97.29 & 70.91  \\ \hline
          \multirow{4}{*}{MobileNets-v1~\cite{Howard_mobilenets:2017}} &  \multirow{4}{*}{6.05M} & 4 & 6392 & 904  & 95.90 & 97.92 & 96.90 & 80.92 & 95.90 & 68.77  \tabularnewline 
			\cline{3-11}
        ~ & ~ & 8 & 6392 & 904  & 96.69 & 98.36 & 97.52 & 82.53 & 96.69 & 70.92  \tabularnewline 
			\cline{3-11}
        ~ & ~ & 12 & 6392 & 904 & 96.61 & 98.36 & 97.48 & 81.36 & 96.61 & 66.52 \tabularnewline 
			\cline{3-11}
        ~ & ~ & 16 & 6392 & 904 & 96.01 & 98.14 & 97.06 & 80.13 & 96.01 & 66.58  \\ \hline
    \end{tabular}
    \vspace{-0.250cm}
\end{table*}

\begin{figure}[t!]
\begin{center}
\includegraphics[width = 0.45\textwidth]{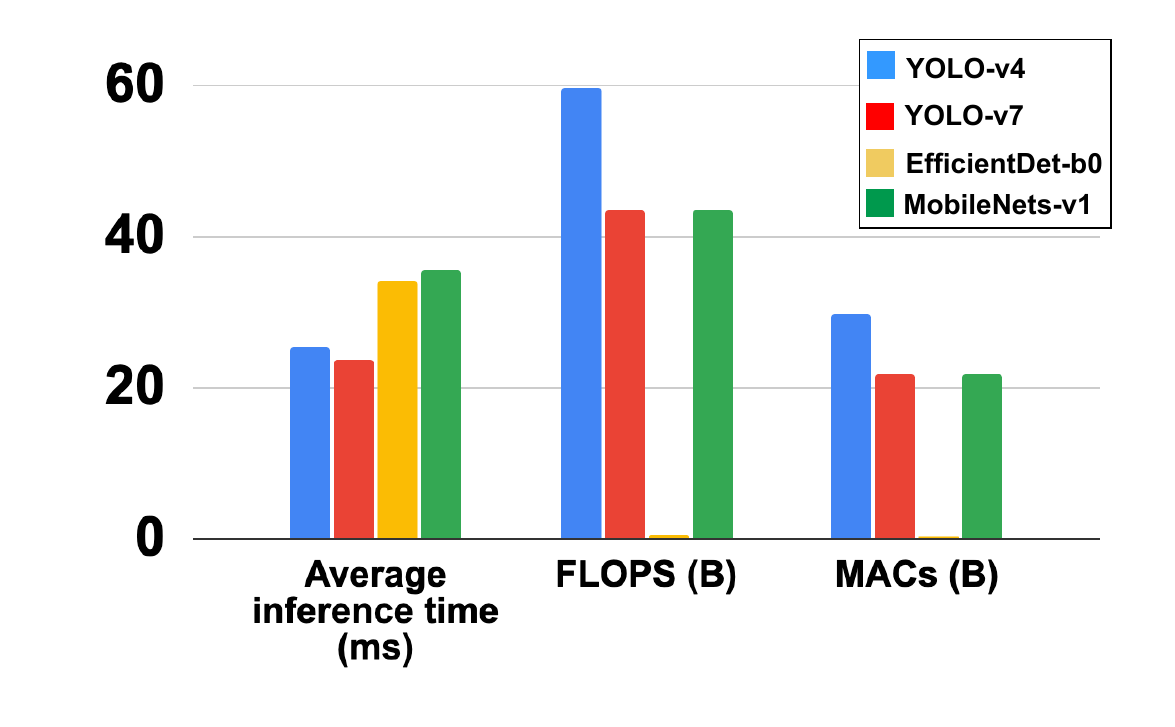}
\vspace{-0.650cm}
\end{center}
\caption[ ]
{YOLO-v4 has the longest inference time and highest FLOPS/ MACs demand, requiring 30.83\% more time than EfficientDet-b0 and 27.99\% more than MobileNets-v1. In contrast, MobileNets-v1 has the shortest inference time and the lowest computational demand.
}
\label{fig:benchmarks_infe_flops_macs}
\vspace{-0.450cm}
\end{figure}

The average inference time was derived using 10 trials each involving 100 unique image inferences. The YOLO-v4 requires 30.83\% and 27.99\% more inference time compared to EfficientDet-b0 and MobileNets-v1 models, respectively, as shown in Figure~\ref{fig:benchmarks_infe_flops_macs}. The YOLO-v4 has $114.70\times$ and $1.37\times$ more FLOPS/MACs compared to EfficientDet-b0 and MobileNets-v1 models, respectively, as shown in Figure~\ref{fig:benchmarks_infe_flops_macs}.

\vspace{-0.350cm}

\section*{Source Code and Scripts}

In this study, we utilized the Aff-Wild2 video dataset~\cite{Kollias_affw2:2018} and conducted analysis using open-source SOTA NN models. 
The dataset analysis scripts are available at the following GitHub repository: \url{https://github.com/riaduli/Thresholded_event_vision_face_dataset}.
%

\section*{Acknowledgment and Interests}
Conceptualization, R. Islam and S.R.S.K. Tummala; 
dataset and analysis, R. Islam, J. Mulé, R. Kankipati, R. Robucci, S. Jalapally, and S.R.S.K Tummala;
original draft preparation, R. Islam;
Review and editing, R. Robucci, C. Howard, and D. Challagundla; funding acquisition, R.Islam and R. Robucci.

This research was funded by the National Science Foundation (NSF) under award number 2138253, the Maryland Industrial Partnerships (MIPS) program under award number MIPS0012, and the UMBC Startup grant. Special thanks to Raiyan Zaman and Rishi Mulchandani for assisting in annotating part of the dataset.

The article authors have declared no conflicts of interest.

\bibliographystyle{IEEEtran}
\bibliography{main}

\begin{thebibliography}{10}
\providecommand{\url}[1]{#1}
\csname url@samestyle\endcsname
\providecommand{\newblock}{\relax}
\providecommand{\bibinfo}[2]{#2}
\providecommand{\BIBentrySTDinterwordspacing}{\spaceskip=0pt\relax}
\providecommand{\BIBentryALTinterwordstretchfactor}{4}
\providecommand{\BIBentryALTinterwordspacing}{\spaceskip=\fontdimen2\font plus
\BIBentryALTinterwordstretchfactor\fontdimen3\font minus
  \fontdimen4\font\relax}
\providecommand{\BIBforeignlanguage}[2]{{%
\expandafter\ifx\csname l@#1\endcsname\relax
\typeout{** WARNING: IEEEtran.bst: No hyphenation pattern has been}%
\typeout{** loaded for the language `#1'. Using the pattern for}%
\typeout{** the default language instead.}%
\else
\language=\csname l@#1\endcsname
\fi
#2}}
\providecommand{\BIBdecl}{\relax}
\BIBdecl

\bibitem{Padilla_evaluation:2012}
R.~Padilla, C.~Costa~Filho, and M.~Costa, ``Evaluation of haar cascade
  classifiers designed for face detection,'' \emph{World Academy of Science,
  Engineering and Technology}, vol.~64, pp. 362--365, 2012.

\bibitem{Wenzhen:2016}
W.~Yuan and S.~Ramalingam, ``Fast localization and tracking using event
  sensors,'' in \emph{2016 IEEE International Conference on Robotics and
  Automation (ICRA)}, 2016, pp. 4564--4571.

\bibitem{Amir_dvs:2017}
A.~Amir, B.~Taba, D.~Berg, T.~Melano, J.~McKinstry, C.~Di~Nolfo, T.~Nayak,
  A.~Andreopoulos, G.~Garreau, M.~Mendoza, J.~Kusnitz, M.~Debole, S.~Esser,
  T.~Delbruck, M.~Flickner, and D.~Modha, ``A low power, fully event-based
  gesture recognition system,'' in \emph{IEEE Conference on Computer Vision and
  Pattern Recognition (CVPR)}, 2017, pp. 7388--7397.

\bibitem{Gallego:2022}
G.~Gallego, T.~Delbrück, G.~Orchard, C.~Bartolozzi, B.~Taba, A.~Censi,
  S.~Leutenegger, A.~J. Davison, J.~Conradt, K.~Daniilidis, and D.~Scaramuzza,
  ``Event-based vision: A survey,'' \emph{IEEE Transactions on Pattern Analysis
  and Machine Intelligence}, vol.~44, no.~1, pp. 154--180, 2022.

\bibitem{Mueggler:2018}
E.~Mueggler, G.~Gallego, H.~Rebecq, and D.~Scaramuzza, ``Continuous-time
  visual-inertial odometry for event cameras,'' \emph{IEEE Transactions on
  Robotics}, vol.~PP, pp. 1--16, 08 2018.

\bibitem{Islam_ncfet:2021}
R.~Islam, ``Negative capacitance clock distribution,'' \emph{IEEE Transactions
  on Emerging Topics in Computing}, vol.~9, no.~1, pp. 547--553, 2021.

\bibitem{Hyeon:2022}
J.-S. Hyeon, S.-H. Kim, and H.-J. Kim, ``A low-power {CMOS} image sensor with
  multiple-column-parallel readout structure,'' \emph{IEEE Journal of the
  Electron Devices Society}, vol.~10, pp. 180--187, 2022.

\bibitem{Islam:2018}
R.~Islam, H.~A. Fahmy, P.~Y. Lin, and M.~R. Guthaus, ``D{CMCS}: Highly robust
  low-power differential current-mode clocking and synthesis,'' \emph{IEEE
  Transactions on Very Large Scale Integration (VLSI) Systems}, vol.~26,
  no.~10, pp. 2108--2117, 2018.

\bibitem{Park:2020}
K.~Park, S.~Yeom, and S.~Y. Kim, ``Ultra-low power {CMOS} image sensor with
  two-step logical shift algorithm-based correlated double sampling scheme,''
  \emph{IEEE Transactions on Circuits and Systems I: Regular Papers}, vol.~67,
  no.~11, pp. 3718--3727, 2020.

\bibitem{Islam_sram:2021}
R.~Islam, B.~Saha, and I.~Bezzam, ``Resonant energy recycling {SRAM}
  architecture,'' \emph{IEEE Transactions on Circuits and Systems II: Express
  Briefs}, vol.~68, no.~4, pp. 1383--1387, 2021.

\bibitem{Guillermo:2019}
\BIBentryALTinterwordspacing
G.~Gallego, T.~Delbr{\"{u}}ck, G.~Orchard, C.~Bartolozzi, B.~Taba, A.~Censi,
  S.~Leutenegger, A.~J. Davison, J.~Conradt, K.~Daniilidis, and D.~Scaramuzza,
  ``Event-based vision: {A} survey,'' \emph{CoRR}, vol. abs/1904.08405, 2019.
  [Online]. Available: \url{http://arxiv.org/abs/1904.08405}
\BIBentrySTDinterwordspacing

\bibitem{Orchard:2015}
\BIBentryALTinterwordspacing
G.~Orchard, A.~Jayawant, G.~K. Cohen, and N.~Thakor, ``Converting static image
  datasets to spiking neuromorphic datasets using saccades,'' \emph{Frontiers
  in Neuroscience}, vol.~9, 2015. [Online]. Available:
  \url{https://www.frontiersin.org/articles/10.3389/fnins.2015.00437}
\BIBentrySTDinterwordspacing

\bibitem{Yang_wider:2016}
S.~Yang, P.~Luo, C.~C. Loy, and X.~Tang, ``W{IDER FACE}: {A} face detection
  benchmark,'' in \emph{IEEE Conference on Computer Vision and Pattern
  Recognition (CVPR)}, 2016.

\bibitem{Liu_faceattributes:2015}
Z.~Liu, P.~Luo, X.~Wang, and X.~Tang, ``Deep learning face attributes in the
  wild,'' in \emph{Proceedings of International Conference on Computer Vision
  (ICCV)}, December 2015.

\bibitem{Jain:2010}
V.~Jain and E.~Learned-Miller, ``F{DDB}: {A} benchmark for face detection in
  unconstrained settings,'' University of Massachusetts, Amherst, Tech. Rep.
  UM-CS-2010-009, 2010.

\bibitem{Serrano-gotarredona_linares-barranco:2015}
T.~Serrano-Gotarredona and B.~Linares-Barranco, ``Poker-{DVS} and {MNIST-DVS}.
  their history, how they were made, and other details,'' \emph{Frontiers in
  Neuroscience}, vol.~9, Dec 2015.

\bibitem{Elias_dataset:2017}
E.~Mueggler, H.~Rebecq, G.~Gallego, T.~Delbruck, and D.~Scaramuzza, ``The
  event-camera dataset and simulator: Event-based data for pose estimation,
  visual odometry, and {SLAM},'' \emph{The International Journal of Robotics
  Research}, vol.~36, no.~2, pp. 142--149, 2017.

\bibitem{Lichtsteiner:2008}
P.~Lichtsteiner, C.~Posch, and T.~Delbruck, ``A 128$\times$ 128 120 db 15
  $\mu$s latency asynchronous temporal contrast vision sensor,'' \emph{IEEE
  Journal of Solid-State Circuits}, vol.~43, no.~2, pp. 566--576, 2008.

\bibitem{Rizk:2012}
C.~Rizk, J.~Lin, S.~Kennerly, P.~Pouliquen, A.~Goldberg, and A.~Andreou,
  ``High-performance, event-driven, low-cost, and swap imaging sensor for
  hostile fire detection, homeland protection, and border security,''
  \emph{Proceedings of SPIE - The International Society for Optical
  Engineering}, vol. 8359, pp. 18--, 05 2012.

\bibitem{Joseph_fris:2012}
\BIBentryALTinterwordspacing
J.~H. Lin, P.~O. Pouliquen, A.~G. Andreou, A.~C. Goldberg, and C.~G. Rizk,
  ``{Flexible readout and integration sensor {(FRIS)}: a bio-inspired,
  system-on-chip, event-based readout architecture},'' in \emph{Infrared
  Technology and Applications XXXVIII}, B.~F. Andresen, G.~F. Fulop, and P.~R.
  Norton, Eds., vol. 8353, International Society for Optics and
  Photonics.\hskip 1em plus 0.5em minus 0.4em\relax SPIE, 2012, p. 83531N.
  [Online]. Available: \url{https://doi.org/10.1117/12.919584}
\BIBentrySTDinterwordspacing

\bibitem{Niwa_isscc:2023}
A.~Niwa, F.~Mochizuki, R.~Berner, T.~Maruyarma, T.~Terano, K.~Takamiya,
  Y.~Kimura, K.~Mizoguchi, T.~Miyazaki, S.~Kaizu, H.~Takahashi, A.~Suzuki,
  C.~Brandli, H.~Wakabayashi, and Y.~Oike, ``A 2.97$\mu m$-pitch event-based
  vision sensor with shared pixel front-end circuitry and low-noise intensity
  readout mode,'' in \emph{IEEE International Solid-State Circuits Conference
  (ISSCC)}, 2023, pp. 4--6.

\bibitem{Liu_adaptive:2018}
M.~Liu and T.~Delbruck, ``Adaptive time-slice block-matching optical flow
  algorithm for dynamic vision sensors.''\hskip 1em plus 0.5em minus
  0.4em\relax BMVC, 2018.

\bibitem{Charbel:2015}
\BIBentryALTinterwordspacing
C.~G. Rizk, J.~P. Wilson, and P.~Pouliquen, ``{Advanced computational sensors
  technology: testing and evaluation in visible, {SWIR}, and {LWIR} imaging},''
  in \emph{Image Sensing Technologies: Materials, Devices, Systems, and
  Applications II}, N.~K. Dhar and A.~K. Dutta, Eds., vol. 9481, International
  Society for Optics and Photonics.\hskip 1em plus 0.5em minus 0.4em\relax
  SPIE, 2015, p. 94810E. [Online]. Available:
  \url{https://doi.org/10.1117/12.2177350}
\BIBentrySTDinterwordspacing

\bibitem{Klenk_Enerf:2023}
S.~Klenk, L.~Koestler, D.~Scaramuzza, and D.~Cremers, ``E-{N}e{RF}: Neural
  radiance fields from a moving event camera,'' \emph{IEEE Robotics and
  Automation Letters}, vol.~8, no.~3, pp. 1587--1594, 2023.

\bibitem{Rudnev_cvpr:2023}
V.~Rudnev, M.~Elgharib, C.~Theobalt, and V.~Golyanik, ``Event{N}e{RF}: Neural
  radiance fields from a single colour event camera,'' in \emph{Proceedings of
  the IEEE/CVF Conference on Computer Vision and Pattern Recognition (CVPR)},
  June 2023, pp. 4992--5002.

\bibitem{Liu_iscas:2017}
M.~Liu and T.~Delbruck, ``Block-matching optical flow for dynamic vision
  sensors: Algorithm and fpga implementation,'' in \emph{IEEE International
  Symposium on Circuits and Systems (ISCAS)}, 2017, pp. 1--4.

\bibitem{Rebecq_esim:2018}
H.~Rebecq, D.~Gehrig, and D.~Scaramuzza, ``Esim: an open event camera
  simulator,'' in \emph{Conference on robot learning}.\hskip 1em plus 0.5em
  minus 0.4em\relax PMLR, 2018, pp. 969--982.

\bibitem{Hu_v2e:2021}
\BIBentryALTinterwordspacing
Y.~Hu, S.~C. Liu, and T.~Delbruck, ``v2e: From video frames to realistic {DVS}
  events,'' in \emph{2021 {IEEE/CVF} Conference on Computer Vision and Pattern
  Recognition Workshops ({CVPRW})}.\hskip 1em plus 0.5em minus 0.4em\relax
  IEEE, 2021. [Online]. Available: \url{http://arxiv.org/abs/2006.07722}
\BIBentrySTDinterwordspacing

\bibitem{Sengupta:2022}
\BIBentryALTinterwordspacing
J.~Sengupta, S.~Liu, and A.~Andreou, ``Retinosim: An event-based data synthesis
  tool for neuromorphic vision architecture exploration,'' in \emph{Proceedings
  of the International Conference on Neuromorphic Systems 2022}, ser. ICONS
  '22.\hskip 1em plus 0.5em minus 0.4em\relax New York, NY, USA: Association
  for Computing Machinery, 2022. [Online]. Available:
  \url{https://doi.org/10.1145/3546790.3546805}
\BIBentrySTDinterwordspacing

\bibitem{Venkatachalam:2023}
S.~Venkatachalam, V.~S. Vivekanand, and R.~Kubendran, ``Frame of events: A
  low-latency resource-efficient approach for stereo depth maps,'' in
  \emph{International Conference on Automation, Robotics and Applications
  (ICARA)}, 2023, pp. 324--328.

\bibitem{Redmon_yolo:2016}
J.~Redmon, S.~Divvala, R.~Girshick, and A.~Farhadi, ``You only look once:
  Unified, real-time object detection,'' 2016.

\bibitem{Bochkovskiy_yolov4:2020}
A.~Bochkovskiy, C.-Y. Wang, and H.-Y.~M. Liao, ``Y{OLO}v4: Optimal speed and
  accuracy of object detection,'' 2020.

\bibitem{Wang_yolov7:2022}
C.-Y. Wang, A.~Bochkovskiy, and H.-Y.~M. Liao, ``{YOLOv7: Trainable
  bag-of-freebies sets new state-of-the-art for real-time object detectors},''
  2022.

\bibitem{Wang_cspnet:2020}
C.-Y. Wang, H.-Y.~M. Liao, Y.-H. Wu, P.-Y. Chen, J.-W. Hsieh, and I.-H. Yeh,
  ``C{SPN}et: A new backbone that can enhance learning capability of {CNN},''
  in \emph{Proceedings of the IEEE/CVF conference on computer vision and
  pattern recognition workshops}, 2020, pp. 390--391.

\bibitem{Tan_efficientdet:2020}
M.~Tan, R.~Pang, and Q.~V. Le, ``Efficient{D}et: Scalable and efficient object
  detection,'' 2020.

\bibitem{Howard_mobilenets:2017}
A.~G. Howard, M.~Zhu, B.~Chen, D.~Kalenichenko, W.~Wang, T.~Weyand,
  M.~Andreetto, and H.~Adam, ``Mobile{N}ets: Efficient convolutional neural
  networks for mobile vision applications,'' 2017.

\bibitem{Kollias_affw2:2018}
\BIBentryALTinterwordspacing
D.~Kollias and S.~Zafeiriou, ``Aff-wild2: Extending the aff-wild database for
  affect recognition,'' \emph{CoRR}, vol. abs/1811.07770, 2018. [Online].
  Available: \url{http://arxiv.org/abs/1811.07770}
\BIBentrySTDinterwordspacing

\end{thebibliography}

\end{document}